\documentclass[a4]{sab}
\usepackage{graphicx}

\title{Detecting Novel Features of an Environment Using Habituation \footnote{In From Animals to Animats, Proceedings of the Sixth International Conference on Simulation of Adaptive Behaviour, 2000}}
\author{Stephen Marsland \and Ulrich Nehmzow \and Jonathan Shapiro}
\affiliation{Department of Computer Science\\University of Manchester\\Oxford Road\\Manchester M13 9PL\\
\texttt{\{smarsland, ulrich, jls\}@cs.man.ac.uk}}

\begin{document}
\maketitle

\begin{abstract}

In this paper a novelty filter is introduced which allows a robot operating in an unstructured
environment to produce a self-organised model of its surroundings and to
detect deviations from the learned model. The environment is perceived using the
robot's 16 sonar sensors. The algorithm produces a novelty measure for each sensor scan 
relative to the model it has learned. This means that it highlights stimuli which have
not been previously experienced. The novelty filter proposed uses a model of habituation.
Habituation is a decrement in behavioural response when a stimulus is presented repeatedly.
Robot experiments are presented which demonstrate 
the reliable operation of the filter in a number of environments.

\end{abstract}

\section{Introduction}

The ability to detect and respond suitably to novel stimuli, that is
new or changed features within an environment, is very useful
to animals and robots. Studies have shown that animals 
can rapidly recognise changes in their environment~\cite{OKeefe77,Knight96}.
This ability has two main purposes - to avoid predators and to find potential 
food~\cite{Pribram92}. 
This paper describes an algorithm suitable for detecting novel stimuli 
and applies it to an autonomous
agent. The filter learns to recognise features which it has seen
before and evaluates the novelty of sonar scans taken as a robot 
explores an environment. 

Experiments are presented which demonstrate that the algorithm
can learn an internal representation of {\em one} environment
and then use this model in a {\em second}, similar, environment to detect
new features which were not present in the first environment. 
The system could therefore be used in inspection tasks, being trained in
a section of the environment that is known to be `clean', i.e., containing
no undesired features. Having learned in the `clean' environment, the robot could
move into the wider area, highlighting those features which were novel. 
These would be features which were not present in the original environment.
A novelty filter can also be used to direct the attention of the
robot to new stimuli, so that the amount of processing needed
to deal with its sensory perceptions can be reduced. It can also
be used to decide when a feature should be learned by recognising that
it is new, that is, different to previous perceptions.

\subsection{Habituation}

Habituation -- a reduction in behavioural response when a stimulus is
perceived repeatedly -- is thought to be a fundamental mechanism for
adaptive behaviour. It is present in animals from the sea slug
{\em Aplysia}~\cite{Bailey83,Greenberg87} to humans~\cite{OKeefe77}
through toads~\cite{Ewert78,Wang92} and cats~\cite{Thompson86}. In
contrast to other forms of behavioural decrement, such as fatigue,
a change in the nature of the stimulus restores the response to its original level -
a process known as dishabituation. In addition, if a particular
stimulus is not presented for a period of time, the response is
recovered, a form of `forgetting'. An overview of the effects and
causes of habituation can be found in~\cite{Thompson66,Peeke73}.

\begin{figure}
\centering
\includegraphics[angle=270,width=.45\textwidth]{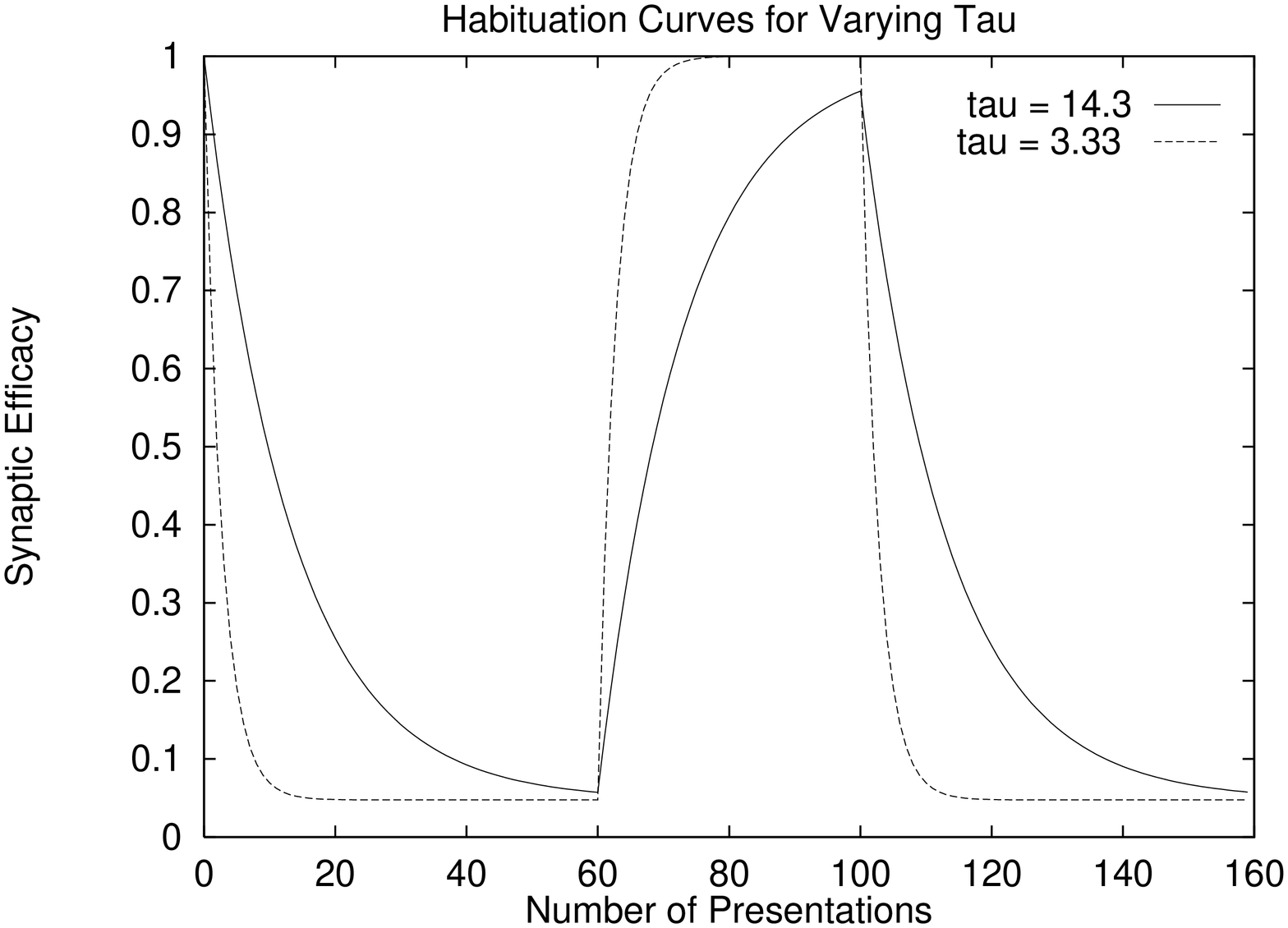}
\caption{
\textsf {\small An example of how the synaptic efficacy drops when habituation occurs. 
In both curves, a constant stimulus $S(t)=1$ is presented,
causing the efficacy to fall. The stimulus is reduced to $S(t)=0$
at time $t=60$ where the graphs rise again, and becomes
$S(t)=1$ again at $t=100$, causing another drop. 
The two curves show the effects of varying $\tau$ in equation~\ref{HabEqn}.
It can be seen that a larger value of $\tau$ causes both the learning
and forgetting to occur faster. The other variables were the same for
both curves, $\alpha = 1.05$ and $y_0 = 1.0$.}}
\label{curves}
\end{figure}

A number of researchers have produced mathematical models of the
effects of habituation on the efficacy of a synapse. They include
%Groves and Thompson~\cite{Groves70}, Stanley~\cite{Stanley76} and Wang and Hsu~\cite{Wang90}.
Groves and Thompson, Stanley, and Wang and Hsu~\cite{Groves70,Stanley76,Wang90}.
The models are similar, except that Wang and Hsu's
allows for long term memory, while the others do not. Long term memory means that
an animal habituates more quickly to a stimulus to which it
has habituated previously.
For simplicity, the model which is used in the work presented here is that of
Stanley. In his model the synaptic efficacy, $y(t)$, decreases according to the
following equation:

\begin{equation}
 \tau \frac{dy(t)}{dt} = \alpha \left[ y_0 - y(t) \right] - S(t),
\label{HabEqn}
\end{equation}

\noindent
where $y_0$ is the original value of $y$, $\tau$ and $\alpha$
are time constants governing the rate of habituation and 
recovery respectively, and $S$ is the stimulus presented.
A graph of the effects of the equation can be seen in figure~\ref{curves}.

\subsection{Novelty Detection}

Habituation can be used to detect novel stimuli in a very simple
way. A series of habituating neurons can be used to build an internal
representation of perceived stimuli, and then stimuli not included
in this model will be the only ones which reach the attention of the
animal. It has been suggested~\cite{Heiligenberg80, Grau86} that 
pulse-type weakly electric fish use a mechanism similar to this
when they use the returns from weak electric pulses to sample their
environment. Heiligenberg~\cite{Heiligenberg80} suggests that the
fish store a `template' of responses received from their environment,
and detect novelty with respect to this memory, changing the template
if novel stimuli remain for several samples.
%The novelty filter built along these principles is described in section~\ref{HSOM}.

\subsection{Related Work}

A number of novelty detection techniques have been proposed in
the neural network literature. Most of them are trained
off-line. For example, the Kohonen Novelty Filter~\cite{Kohonen76,Kohonen93},
is an autoencoder neural network trained using back-propagation 
of error, so that the network extracts the principal components of the input. 
After training, any input presented to the network
produces one of the learned outputs, and the bitwise
difference between input and output highlights novel
components of the input. Several authors have investigated the
effects of this filter, notably  Aeyels~\cite{Aeyels90} who
added a `forgetting' term into the equations.

Other approaches include that of Ho and Rouat \cite{Ho98},
who used an integrate-and-fire model inspired by layer IV of the cortex.
Their model measures how long an oscillatory network takes to converge
to a stable solution, reasoning that previously seen inputs 
should converge faster than novel ones.

Several researchers have used the gated dipole \cite{Grossberg72a},
a construct which compares current and recent values of a stimulus.
Levine and Prueitt~\cite{Levine92} used it
to compare inputs with pre-defined stimuli, novel inputs
causing greater output values. They proposed a model where a number
of gated dipoles were linked into a `dipole field' and used it
to model novelty preference in frontally lesioned rhesus monkeys. 
The gated dipole has
also been used by \"{O}\v{g}men and Prakash~\cite{Ogmen97}
who used it as part of a system to control a robot arm, which moved its
end effector to places which were novel. This was done by quantising
the workspace and associating a gated dipole with each separate area. Their
work also considered the question of detecting novel objects. They did this
by taking the output of an ART network~\cite{Carpenter88}, which attempts to classify
the objects into a number of categories, and feeding it into a network of gated dipoles.

Ypma and Duin~\cite{Ypma97} proposed a novelty detection mechanism
based on the self-organising map. The distance of the winning neuron
from neighbourhoods which have fired recently was calculated, and those
beyond a certain threshold were counted as novel. This method was used
by Taylor and MacIntyre~\cite{Taylor98} to detect faults when monitoring
machines. The network was trained on data taken from machines operating
normally, and data deviating from this pattern was taken as novel. This is a common 
technique when faced with a problem for which there is very little data
in one class, relative to others. Examples include machine breakdowns~\cite{Nairac99,Worden99}
and mammogram scans~\cite{Tarassenko95}. Often supervised techniques
such as Gaussian Mixture Models or Parzen Windows are used, and the 
problem reduces to attempting to recognise when inputs do not
belong to the distribution which generates the normal data~\cite{Bishop94}.
The method proposed by Ypma and Duin relies 
very strongly on the choice of threshold and on the properties
of the data presented to the network, which must form strictly segmented
neighbourhood clusters without much spread.

Growing networks such as Adaptive Resonance Theory (ART)~\cite{Carpenter88} 
can be used to define as novel those things which have never been seen before,
by using a new, uncommitted node to represent them.
The use of habituation allows novelty to be defined more specifically as those things which have
not been seen in the current context. The filter does this by learning an online,
adaptive representation of the current environment. Habituation also
allows the novelty of a stimulus to be evaluated, so that the novelty reduces
with perception over time. The combination of the two ideas, amalgamating growing
networks with habituation has been considered in~\cite{Marsland00}.

\section{The Habituating Self-Organising Map (HSOM) \label{HSOM}}

\begin{figure}
\centering
\includegraphics[angle=270,width=.45\textwidth]{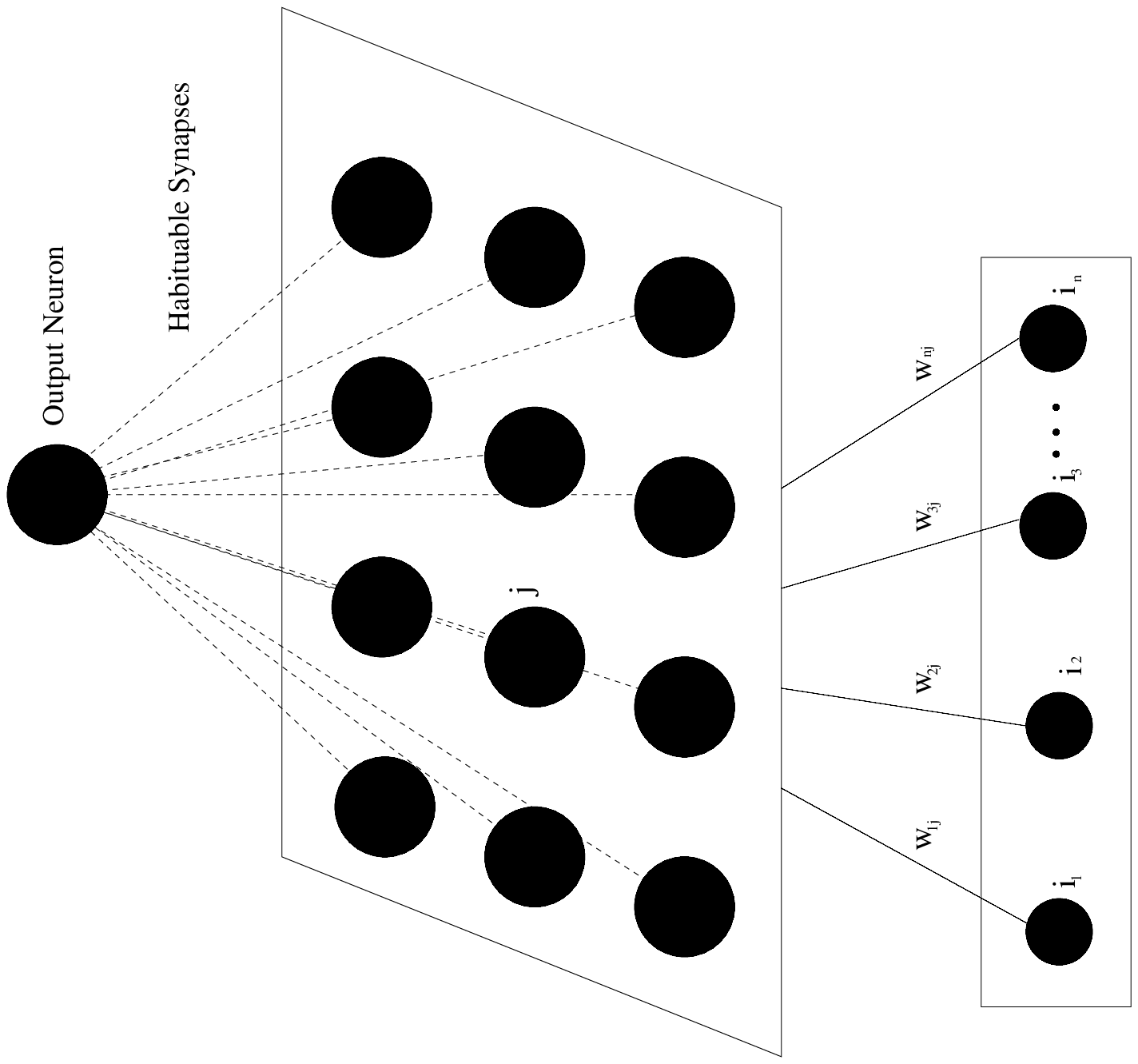}
\caption{ \textsf{ \small {The novelty filter.
The input layer connects to a clustering layer which
represents the feature space, the winning neuron (i.e.,
the one `closest' to the input) passing its
output along a habituable synapse to the output neuron so
that the output received from a neuron reduces with the
number of times it fires.}}}
\label{hsom}
\end{figure}

This section introduces the novelty filter on which this work is based.
An input vector is presented to a clustering network, which finds a winning neuron
using a winner-takes-all strategy.
Each neuron in the map field is connected to the output neuron (see 
figure~\ref{hsom}) via a habituable synapse, so that the 
more frequently a neuron fires, the lower the efficacy of
the synapse and hence the lower the strength of the output. The behaviour
of the habituable synapses is controlled by equation~\ref{HabEqn}.
The strength of the winning synapse is taken as a novelty
value for the particular winning neuron, and hence the perception presented, 
with more novel stimuli having values closer to 1, and more common
stimuli values closer to 0.

There are two choices of how to deal with the output synapses of neurons
which do not belong to the winning neighbourhood. They could be left without
any input, so that they do not habituate and their value remains unchanged,
or, instead, a zero input ($S(t)=0$) could be given. In this case the 
synapses will `forget' previous inhibition over time, with the time
controlled by the constant $\tau$. This can be seen in the second,
ascending part of figure~\ref{curves}.
In the results presented here the network remembers all perceptions,
forgetting is not used and the synapses of neurons which are not in
the winning neighbourhood do not receive any input.

In the implementation described below, a Kohonen Self-Organising 
Map (SOM) implementing Learning Vector Quantisation~\cite{Kohonen93} 
is used as the clustering mechanism. Kohonen networks are often used 
for problems dealing with robot sensory inputs~\cite{Kurz96}. Although 
guaranteeing convergence of a two dimensional network is still an unsolved problem~\cite{Erwin92a},
the network performs a useful dimensionality reduction and clusters
the perceptions in this lower dimensional space. The network calculates
the distance between the input and each of the neurons in
the map field where the distance is defined by:

\begin{equation}
d = \sum_{i=0}^{N-1} \left( \mathbf{w}_i (t) - \mathbf{v} (t) \right) ^2,
\end{equation}

\noindent
where $\mathbf{v} (t)$ is the input vector at time $t$,
$\mathbf{w}_{i}$ the weight between input $i$ and the neuron and
the sum is over the $N$ components of the input vector.
The neuron with the minimum $d$ is selected and the
weights for that neuron and its eight topological neighbours
are updated by:

\begin{equation}
\mathbf{w}_{i} (t+1) = \mathbf{w}_{i} (t) + \eta (t) \left( \mathbf{v} (t) - \mathbf{w}_{i} (t) \right)
\end{equation}

\noindent
where $\eta$ is the learning rate, $0 \leq \eta (t) \leq 1$.
The habituable synapses for both the winning neuron and its
topological neighbours were updated, using equation~\ref{HabEqn} with $S(t)=1$.
The value of $\tau$ used was different for the winner and its neighbours, being $\tau = 3.33$ for the
winner and $\tau = 14.3$ for the neighbourhood synapses. This meant that winning neurons
habituated quickly, while neighbourhood neurons, which recognise similar perceptions,
have a smaller amount of habituation.
The effects of these values can be 
seen in figure~\ref{curves}. The synapses of other neurons were not affected. 
Other variables kept their
values regardless of which neuron is firing, being
$\alpha = 1.05$ and $y_0=1.0$. With these values the
synaptic efficacy never falls below 0.0476 (see figure~\ref{curves}),
a function of the values used for the constants in
equation~\ref{HabEqn}.

A square map field,
comprising 100 neurons arranged in a 10 by 10 grid, was used. 
The neighbourhood size was kept constant at $\pm 1$ unit in all directions
and the learning rate $\eta$ was 0.25, so that the network was always learning.

\section{Experiments}

The experiments presented
investigate the ability of the novelty filter to learn a model of an external environment
through periodic sonar scans taken whilst exploring, and to detect deviations from that model.

\subsection{The Robot}

\begin{figure}
\centering
\includegraphics[width=.45\textwidth]{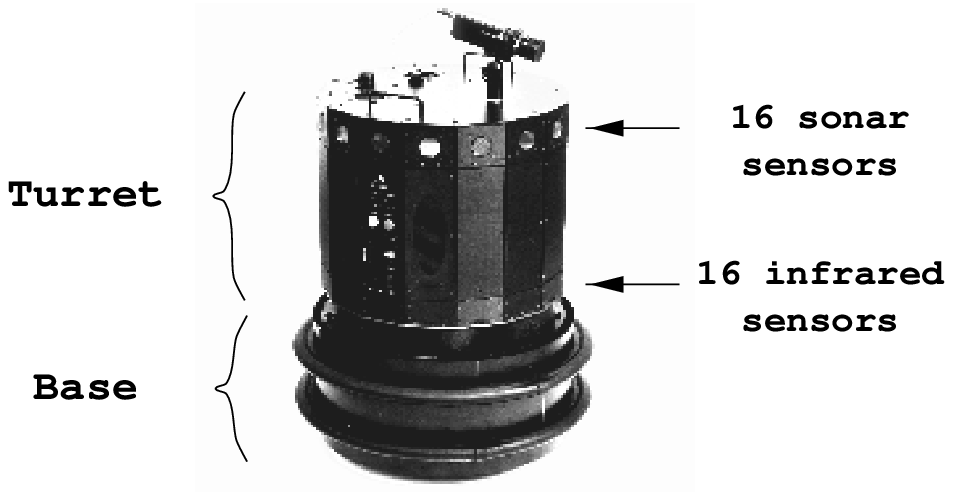}
\caption{ \textsf{ \small {The Nomad 200 mobile robot.}}}
\label{FortyTwo}
\end{figure}

A Nomad~200 mobile robot (shown in figure~\ref{FortyTwo}) was used to perform the experiments. The
band of infra-red sensors mounted at the bottom of the turret of the robot were 
used to perform
a pre-trained wall-following routine~\cite{Nehmzow94a}, and the 
16 sonar sensors at the top of the turret were used to provide perceptions
of the robot's environment. The angle between the turret and base 
of the robot was
kept fixed. 
The input vector to the novelty filter
consisted of the 16 sonar sensors, each normalised to be between 0 and 1 
and thresholded at about 4~metres.
The readings were inverted so that inputs from sonar responses received from closer objects were greater.

\subsection{Description of the Experiments \label{Description}}

\begin{figure*}
\centering
\includegraphics[width=\textwidth,height=.9\textheight]{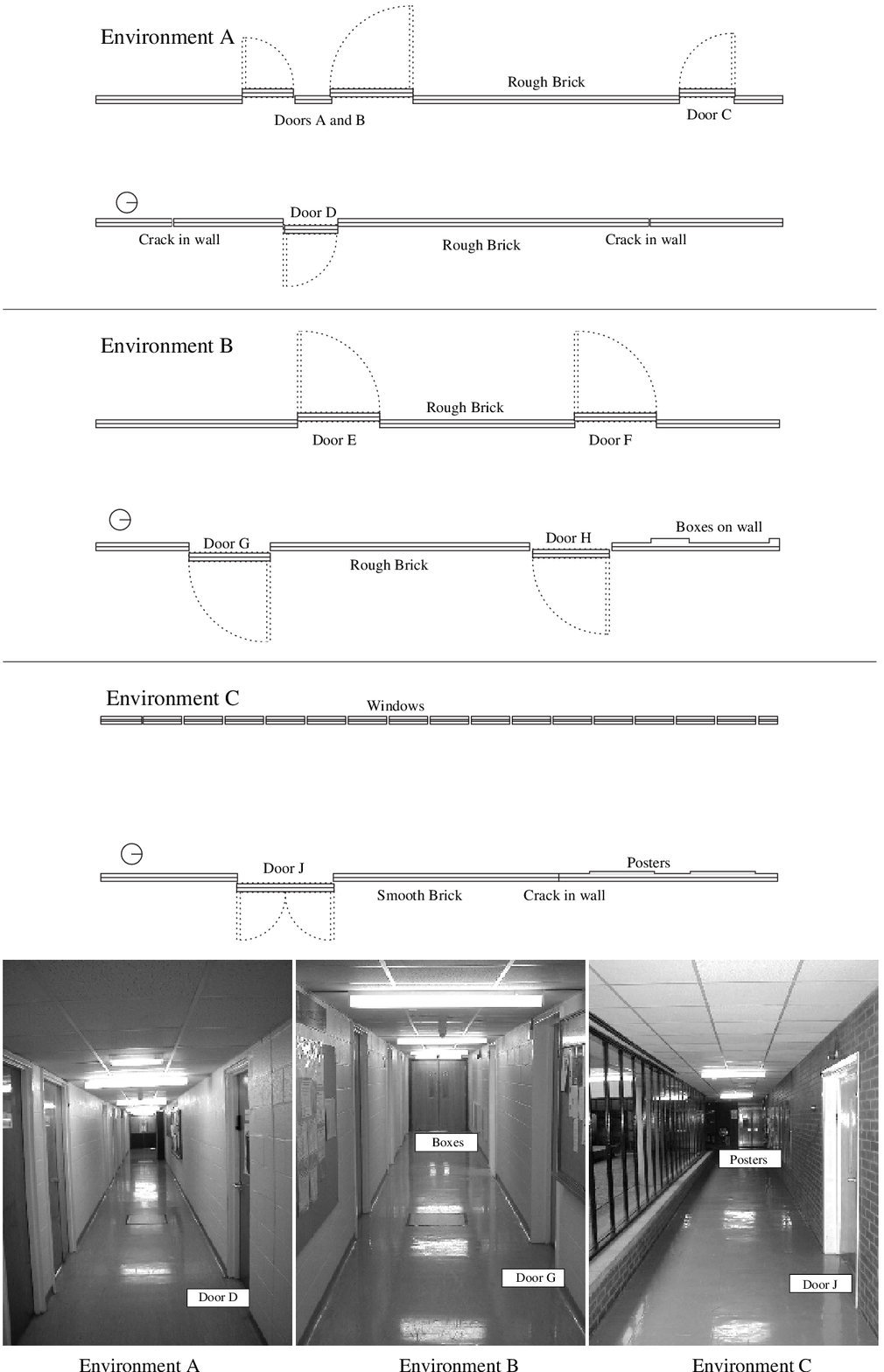}
\caption{ \textsf{ \small {Diagrams of the three environments used.
The robot is shown facing in the direction of travel adjacent to the
wall that it followed. Environments A and B are two similar sections
of corridor, while environment C is wider and has walls made of
different materials, glass and brick instead of breezeblock.
The photographs show the environments as they appear from the
starting position of the robot. The notice boards which are visible
in environment B are above the height of the robot's sonar sensors,
and are therefore not detected.}}}
\label{envs}
\end{figure*}

Each experiment consisted of a number of trials, with each trial
taking the same form. The robot was positioned
at a starting point chosen arbitrarily within the environment. From this starting point 
the robot travelled for 10~metres using a wall--following behaviour, training the
HSOM as sensory stimuli were perceived, and then
stopped and saved the weights of the HSOM. Approximately every 10\,cm
along the route the smoothed values of the sonar perceptions over that 10\,cm
of travel were presented to the novelty filter, which produced a
novelty value for that perception. At the end of the run 
a manual control was used to return the robot to the beginning,
and the same procedure repeated, starting with the weights learned
during the previous run. 

The trials were performed in pairs. After every training run, where 
the sonar readings were used to train the weights of the HSOM, a
second, non-learning, trial was performed. In this run the sonar inputs
generated outputs from the novelty filter, which records how novel
the perceptions were.
At the beginning of the first trial every perception
was novel, as they had not been perceived before, but after a short distance 
the filter began to recognise some similar perceptions, and after
a number of runs in an environment an accurate representation
had been reached so that no perceptions were seen as novel, i.e.,
the output of the novelty filter was only the resting activity of
the output neuron throughout the run.
Once this occurred the environment was changed. This could either
be a modification of the current environment (such as the opening of a door),
or the robot could be moved to a new environment. 
The changes are described in section~\ref{Results}

\subsection{Environments}

Three different environments were used, together with a control 
environment for training, as shown in figure~\ref{envs}.
Environments A and B are two sections of corridor on the
second floor of the Computer Science building at the University
of Manchester. The corridors are 1.7\,m wide and have walls made from
painted breezeblock. Doors made of varnished wood lead from the corridors
into offices.
Environment C is part of the first floor of the building.
It is a section of corridor 2.1\,m wide, with smooth brick on one
side and a wall of glass on the other. This environment
has display boards mounted on the wall, as can be seen in figure~\ref{envs}.
These boards are sufficiently low to be visible to the sonar sensors. They are
made from a laminated, shiny plastic.
In all the experiments the environments were kept static.

\section{Results \label{Results}}

\begin{figure*}
\centering
\vspace{-15mm}
\includegraphics[height=.75\textwidth,width=.5\textheight,angle=90]{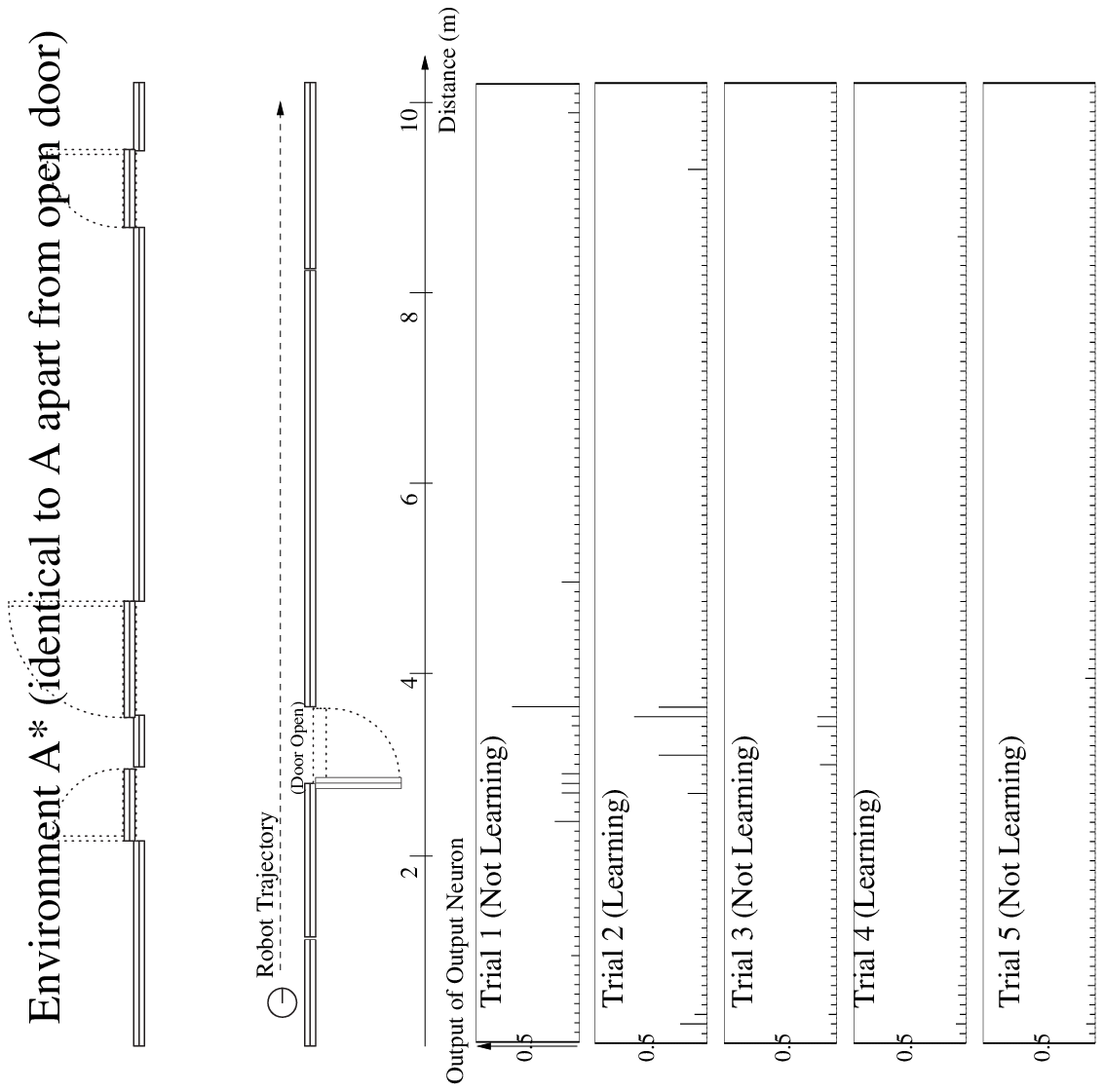} %
\includegraphics[height=.75\textwidth,width=.5\textheight,angle=90]{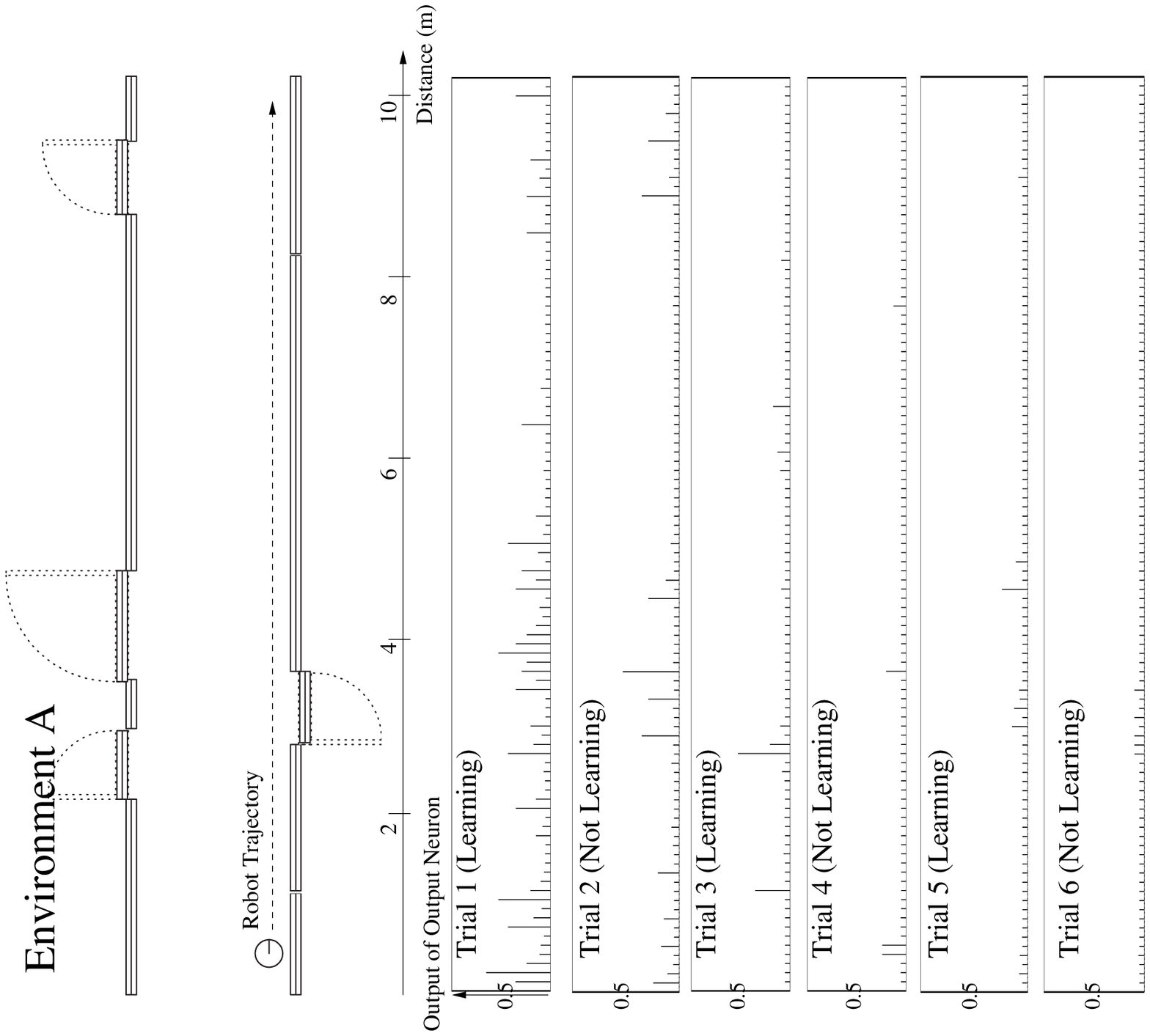} 
\vspace{-17mm}
\caption{ \textsf{ \small {The results of the first experiment. The pictures
on the left show the output of the output neuron of the novelty filter as the robot moves within
environment A when learning and not learning. Once it has stopped detecting
novelty features (so that the activity of the output neuron is small), the
environment was changed by opening a door. The results of this are
shown on the right, and are discussed in section~\ref{Res1}.}}}
\label{Exp1}
\end{figure*}

\subsection{Environment A \label{Res1}}

\begin{table}[hbt]
\begin{center}
\begin{tabular}{|c|l|c|c|}
\hline
% & \multicolumn{5}{|c|}{Second Stimulus} \\
Exp & Trial & \multicolumn{2} {|c|}{Output of the Output Neuron} \\
\cline{3-4}
(Env)& & Integrated &  Maximum \\
\hline
\textbf{1(A)} &&&\\
& 1 (L) & 9.17 & 1.00 \\
& 2 (NL) & 2.64 & 0.53 \\
& 3 (L)& 1.34 & 0.43 \\
& 4 (NL) & 1.07 & 0.42 \\
& 5 (L)& 0.52 & 0.42 \\
& 6 (NL)  & 0.30 & 0.12 \\
\textbf{1(A*)} &&&\\
& 1 (NL)  & 1.44 & 0.65 \\
& 2 (L)& 1.45 & 0.65 \\
& 3 (NL) & 0.37 & 0.17 \\
& 4 (L)& 0.14 & 0.08 \\
& 5 (NL) & 0.01 & 0.05 \\
\textbf{2(B)} &&&\\
& 1 (NL) & 3.38 & 0.65 \\
& 2 (L)& 1.18 & 0.42 \\
& 3 (NL) & 1.21 & 0.42 \\
& 4 (L)& 0.87 & 0.35 \\
& 5 (NL) & 0.03 & 0.08 \\
& Control & 27.09 & 1.00 \\
\textbf{3(C)} &&&\\
& 1 (NL) & 9.68 & 0.65 \\
& 2 (L)& 2.56 & 0.65 \\
& 3 (NL) & 3.06 & 0.53 \\
& 4 (L)& 0.98 & 0.42 \\
& 5 (NL) & 0.29 & 0.15 \\
& Control & 27.75 & 1.00 \\
\hline
\end{tabular}
\end{center}
\caption{\textsf{\small{ The table shows the integrated output
of the output neuron over each trial and the maximum output of the neuron.
The integrated output has had the resting activity of the neuron (0.0476 for each time step)
subtracted. It can be seen that both the integrated output and maximum output
drop after each learning trial (labelled L).}}}
\label{nov}
\end{table}

In the first
experiment, shown in figure~\ref{Exp1}, the robot learned a
representation of environment A (see figure~\ref{envs}), using the procedure
described in section~\ref{Description}, so that it no longer
detected any novelty when traversing it by following the wall. 
The left side of figure~\ref{Exp1} shows the sequence of trials in which the
robot learned an internal
representation of environment A. In the first trial
the novelty filter is learning, as can be seen from the characteristic
habituation shape of successive readings near the beginning of the run
and where the door on the right of the robot is perceived (at about 3 metres).
It can be seen that everything is initially novel, but the filter
rapidly habituates to the brick wall.
In all the experiments the robot gets a much clearer picture of the
right hand side of the environment because that is the wall that it
was following.
The figures show the 
strength of output from the output neuron of the HSOM at each step
along the 10 metre section of corridor which make up each environment.

In the second trial, where the filter is not learning, simply using the weights
learned during the first, there is much less novelty. It is interesting to
note that the crack in the wall near the beginning of the environment is
perceived during the first and third trial, but not the second. This is 
because the crack is very thin and the perceptions of the robot depend
on the speed at which it moves and its precise position. The features
which take longer to learn are those which are perceived less frequently.
For instance, the robot rapidly learns to recognise the perception of walls,
but doorways, which are seen only occasionally, and from many different
angles, take longer. This demonstrates that novel features are those
which have been seen only infrequently, if at all.

In trial 6 it can be seen that the filter has learned a representation of
the environment, because the output of the filter does not rise above
the baseline resting activity of the output neuron. At this point
the environment was altered by opening a door so that the sonar perception
changed (shown as Environment A* on the right hand side of figure~\ref{Exp1}).
In order to prevent
the robot from going through the door a cardboard box was placed in the doorway. This was
of sufficient height to appear on the infra--red sensors,
which were used for wall--following, but did not appear on the
sonar scans. The novelty filter detects the open door, as can be seen from the
burst of novelty in the figure, but successive trials show that the
filter quickly learns to recognise this new feature. No other novel features were detected.
It can be seen that the 
doorways are detected slightly before the robot reaches them and still detected
for a small distance afterwards This is because the sonar scans project in front
and behind of the robot.

\begin{figure*}
\centering
\vspace{-15mm}
\includegraphics[height=.75\textwidth,width=.5\textheight,angle=90]{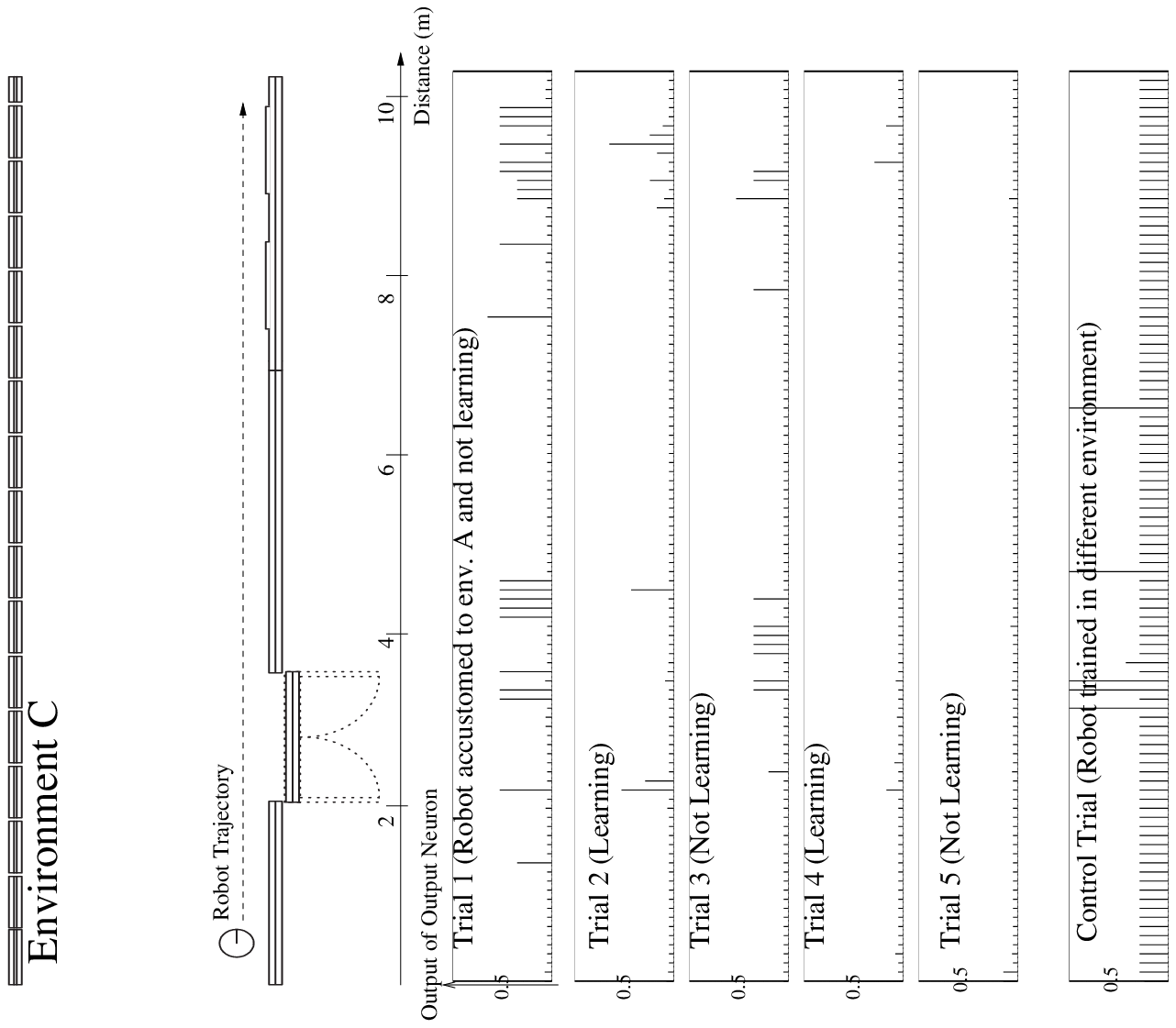} %
\includegraphics[height=.75\textwidth,width=.5\textheight,angle=90]{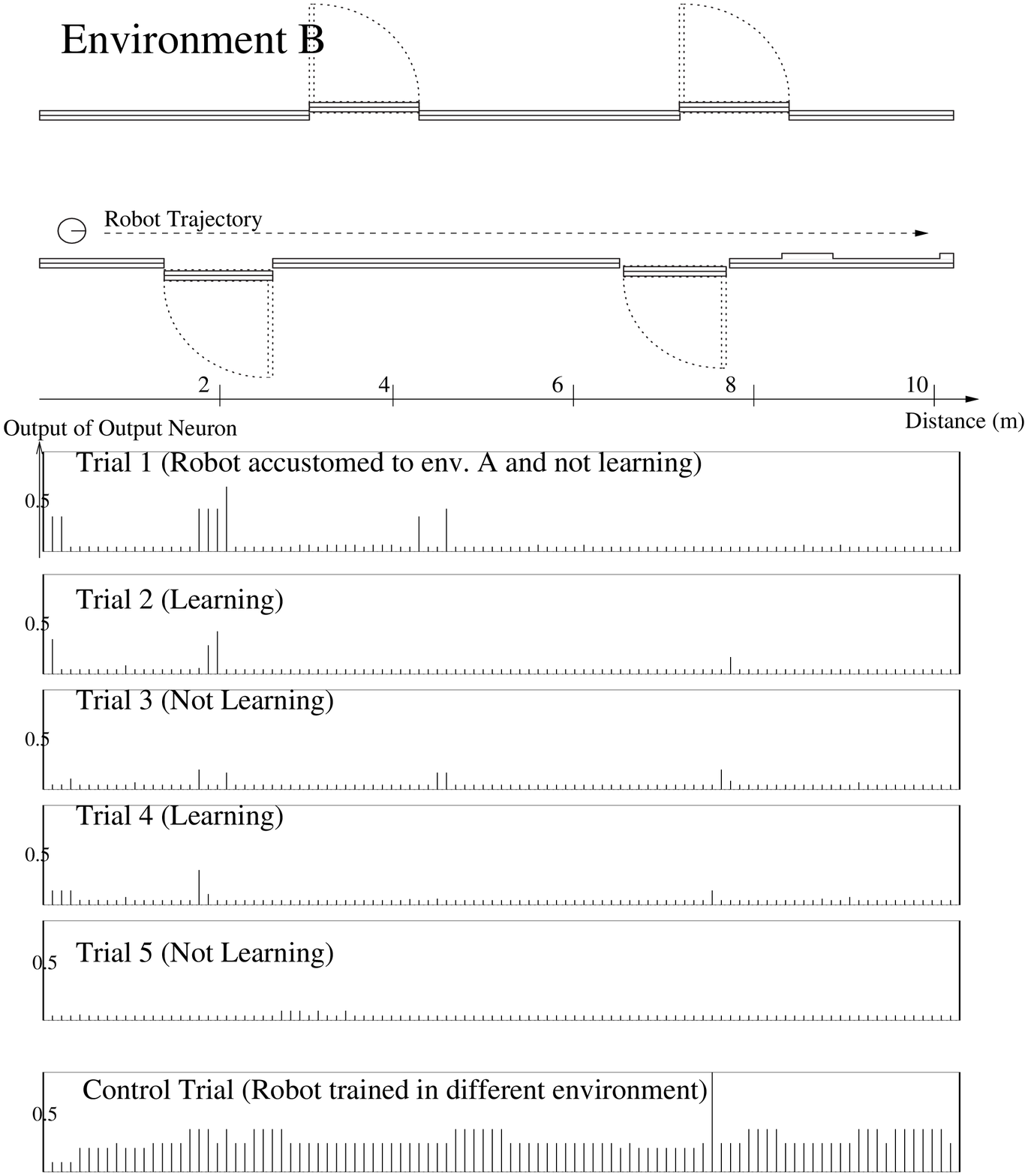} 
\vspace{-15mm}
\caption{ \textsf{ \small {The results of the {\em (left)} second and {\em (right)}
third experiments. Similarly to figure~\ref{Exp1}, they show the output of
the novelty filter to the robot's perceptions. The controls at the bottom of
the figures shown the response to the environments after prior training in a
different environment, which is described in section~\ref{Res2}.}}}
\label{Exp2}
\end{figure*}

\subsection{Environment B \label{Res2}}

For the second experiment, shown on the left hand side of figure~\ref{Exp2}, 
the robot was put into environment B (shown in figure~\ref{envs})
and the novelty filter was initialised with the
weights gained after the representation of environment A had been
learned.
Environment B was a very similar corridor
environment to environment A. The procedure of the first experiment was
repeated, with the robot exploring a 10 metre section of the environment and then
stopping. By comparing the integrated outputs of the output neuron in table~\ref{nov} 
in environments A and B and the left hand side of figures~\ref{Exp1} and \ref{Exp2}, it can be seen that much less
novelty was detected in the second environment than in the first. It can be seen
that during the first trial in environment B, shown on the left of 
figure~\ref{Exp2}, the only place where the output of the neuron is especially high 
(signifying novelty) is
at the first of the two doors on the right hand side of the robot (door G in
figure~\ref{envs}). On inspection of the environment, it was discovered that
this doorway was inset deeper than those in the first corridor. 

The bottom graph on the left hand side of figure~\ref{Exp2} shows the novelty
values when the robot is put into environment B after training in a different
environment. The environment used for the training was a large, open environment. The robot was
steered about a course so that it moved along close to a wall, turned and moved into
the middle of the environment and then returned to the wall. Both the graph and 
table~\ref{nov} show that much more novelty was detected when the network
trained in this control environment was moved into environment B.

\subsection{Environment C} 

The third experiment took the same form as the second, again starting 
from the weights gained during training in environment A.
Using these weights the robot explored the environment labelled C in 
figure~\ref{envs}, the results of which are shown on the right of figure~\ref{Exp2}.
Table~\ref{nov} shows that much more novelty is detected in environment C
than in environment B when the robot first explores it. This is to be expected
because the environment is significantly different, being wider and having
walls made of different materials (see the pictures at the bottom of figure~\ref{envs}). 
In particular, the doorway, which is particularly deeply inset, causes a high output from the HSOM,
as do the posters at the end of the environment. Again, the control trial
shows that the novelty filter detects some similarity between environments
C and A, which are both corridor environments but not between C and the open area
in which the control filter was trained. The crack in the wall (at about the 7\,m mark),
which is a similar perception to that seen in environment A, was not detected. 

\section{Summary and Conclusions}

A novelty filter has been proposed that uses habituation
to assign a novelty value to perceptions. The filter learns on-line.
The novelty filter has been demonstrated to be
capable of learning a representation of features of an environment
whilst running on-line on a mobile robot. Experimental results show
that the filter can accurately detect novel features of
the environment as deviations from the learned model. The effects of
training the filter in one environment and then testing it in
another show that if the two environments are similar, the filter
accurately recognises the familiar features, and highlights new features.
This is an adaptive behaviour, as both the network and habituation
weights change so that the results of training the filter in one environment
and then using it in another are different to simply using it in the
second environment.

Future work will focus on two areas. The output of additional sensory
systems will be added so that the filter gains more information
about its environment, and alternative clustering mechanisms
will be considered.

The current work showed that the filter works using sonar scans as inputs.
Sonar readings are inherently noisy, but the novelty filter deals with
this problem through the use of the self-organising map.
In order to perceive more information about the environment, it
will be necessary to integrate further sensor modes into the model.
In particular, a monochrome CCD camera will be used. The images
will have to be extensively preprocessed in
order for the robot with its limited processing power to be able to deal
with them sufficiently quickly.
How the filter will deal with images, and how the sonar and
image novelty filters will be integrated is currently under investigation.
The vision system will be useful to give more information about the
environment, and to make possible an analysis of the scene in order to 
clarify precisely which stimuli in a perception are novel.

Other work will investigate alternatives to the Self-Organising
Map as a suitable clustering mechanism. A number of well-documented
problems with the SOM exist for problems like this, in particular 
the fact that the size of the network has to be decided before
it is used and remains fixed. This is a problem because the 
network can become saturated so that 
previously learned perceptions are lost, or novel stimuli are
misclassified as normal. These are aspects of the plasticity--stability
dilemma~\cite{Carpenter88}. Possible solutions include using a growing
network such as the Growing Neural Gas of Fritzke~\cite{Fritzke95}.
An alternative is to consider a Mixture of Experts~\cite{Jordan94}, with each
expert being responsible for a particular type of perception and
voting on its novelty. 
A committee of networks~\cite{Perrone93} could be used in a similar way.

Another interesting problem is how to deal with features which are spread out
over time in the robot's perception. For example, for the robot travelling
down a corridor, a door frame is always followed by a door. If the robot learns
this, then if only one of this pair of perceptions is detected it will be considered
novel. This is a question of temporal learning~\cite{Chappell93,Euliano98}.

\section*{Acknowledgements}

This research is supported by a UK EPSRC Studentship.

\small
\bibliography{thebib,Manbib}
\bibliographystyle{sab}

\end{document}